\def\year{2022}\relax
\documentclass[letterpaper]{article} 
\usepackage{aaai22}  
\usepackage{times}  
\usepackage{helvet}  
\usepackage{courier}  
\usepackage[hyphens]{url}  
\usepackage{graphicx} 
\usepackage{natbib}  
\usepackage{caption} 
\DeclareCaptionStyle{ruled}{labelfont=normalfont,labelsep=colon,strut=off} 
\usepackage{algorithm}
\usepackage{algorithmic}

\usepackage{newfloat}
\usepackage{listings}
\floatstyle{ruled}
\newfloat{listing}{tb}{lst}{}
\floatname{listing}{Listing}
\title{Distributed Machine Learning and the Semblance of Trust}
\author{
    Dmitrii Usynin \textsuperscript{\rm{1} \rm{2} \rm{3}},
    Alexander Ziller \textsuperscript{\rm{2} \rm{3}},
    Daniel Rueckert \textsuperscript{\rm{1} \rm{2}}, 
    Jonathan Passerat-Palmbach \textsuperscript{\rm{1}},
    Georgios Kaissis  \textsuperscript{\rm{1} \rm{2} \rm{3}}
}
\affiliations{
    \textsuperscript{\rm 1}Department of Computing, Imperial College London\\
    \textsuperscript{\rm 2}Institute for Artificial Intelligence and Informatics in Medicine, Technical University of Munich \\
    \textsuperscript{\rm 3}Institute of Diagnostic and Interventional Radiology, Technical University of Munich

}

\usepackage{dirtytalk}
\usepackage{booktabs}
\begin{document}

\maketitle

\begin{abstract}
The utilisation of large and diverse datasets for machine learning (ML) at scale is required to promote scientific insight into many meaningful problems. However, due to data governance regulations such as GDPR as well as ethical concerns, the aggregation of personal and sensitive data is problematic, which prompted the development of alternative strategies such as distributed ML (DML). Techniques such as Federated Learning (FL) allow the data owner to maintain data governance and perform model training locally without having to share their data. FL and related techniques are often described as \textit{privacy-preserving}. We explain why this term is not appropriate and outline the risks associated with over-reliance on protocols that were not designed with formal definitions of privacy in mind. We further provide recommendations and examples on how such algorithms can be augmented to provide guarantees of governance, security, privacy and verifiability for a general ML audience without prior exposure to formal privacy techniques.
\end{abstract}

\section{Introduction}
Machine learning (ML) has shown promise in solving a number of important problems such as disease survival prediction \citep{vanneschi2011comparison, yan2020machine} or early-stage cancer discovery \citep{kourou2015machine}. However, in order to train ML models capable of solving these tasks effectively, as well as fulfilling requirements such as fairness and generalisation, large, high-quality and unbiased datasets are required. Procuring these datasets has been problematic in contexts that rely on scarce, sensitive data especially in the context of healthcare. So far, this issue has been alleviated through data centralisation, where these datasets are aggregated at a single location, and model training is performed. However, the introduction of stricter data protection and governance regulations (e.g. the general data protection regulation \cite{radley-gardner_fundamental_2016}), and an increased societal awareness of privacy issues resulting in public resistance against aggressive data collection, centralisation of data is increasingly becoming an unsustainable solution. 

As a result, the ML community has witnessed a surge in interest for scalable, privacy-preserving technologies allowing large-scale training of ML models on distributed data across geographically distant institutions. The common factor of this family of protocols is their reliance on sharing the algorithms or the model updates instead of the datasets, thus remaining in line with existing data governance and data protection regulations. This allows the researchers to conduct collaborative machine learning at scale while minimising the data transmitted. One such mechanism, namely \textit{federated learning} (FL) \cite{konevcny2016federated}, has received particular appraisal from the research community and has seen widespread adoption in a large number of studies \cite{bonawitz2019towards,nasirigerdeh2020splink,brisimi2018federated,roy2019braintorrent,nasirigerdeh2021hyfed,zhao2018federated,brisimi2018federated}. However, while FL --by design-- only avoids transmitting data and thus preserves data governance, it is often wrongly referred to as \textit{privacy-preserving}. In fact, collaboratively trained models can unintentionally leak information about the sensitive properties of the training population. Attacks on the models trained and transmitted between institutions have, on numerous occasions, been shown to be able to reveal sensitive information \cite{usynin2021adversarial,he2019model,geiping2020inverting,shokri2017membership,zhang2020secret,ziller2021differentially}. Thus, FL and other governance mechanisms alone are insufficient means of privacy protection and require augmentations in the forms of formal privacy enhancing technologies (PETs). 

Beyond privacy, a number of additional considerations arise when data is processed, independent of whether such processing occurs centrally or --as in FL-- in a decentralised manner. The requirements towards such systems are outlined in a theoretical framework named \textit{structured transparency} (ST) \cite{trask2020beyond, kaissis2020secure}, whose terminology we will employ in the current work. These include:
\begin{enumerate}
    \item The ability to train a model collaboratively without revealing the input data to other contributors (\textbf{input privacy});
    \item The protection of private information which can be learned from the results (i.e. the output) of the computation (\textbf{output privacy});
    \item The ability to verify the origin of the computation's result i.e. that the model update is not submitted by an unauthorised party (\textbf{input verification});
    \item The capability to guarantee the correctness of the output and that the processing of inputs is honest (\textbf{output verification}) and
    \item The ability to control data ownership and exercise governance over it (\textbf{flow governance}).
\end{enumerate}

Adhering to these principles (along with others, such as explainability, that lie outside of scope of this work) is fundamental to frameworks that claim to be trustworthy, giving sound guarantees of privacy to the individuals whose sensitive data is utilised, but also of reliability of the training protocol in general.
Beyond FL, (and in part due to its vulnerabilities), a number of other DML frameworks have been introduced, for instance Swarm Learning (SL) \cite{warnat2021swarm}, Split Learning \cite{vepakomma2018split} or gossip learning \cite{hegedHus2019gossip}). These attempt to address some of the aforementioned requirements, but often fail to offer full ST guarantees either. In this article, motivated by the increasing popularity of DML, we aim to equip ML practitioners interested in applying DML techniques or reading related literature with an overview of relevant techniques, so that they can avoid misunderstandings and common pitfalls.

Our contributions can be summarised as follows:
\begin{itemize}
    \item We provide definitions for the commonly used (and misunderstood) terms \textit{governance, privacy, secrecy, security, accountability} and \textit{verification}.
    \item We contextualise these techniques within the structured transparency framework.
    \item Finally, we discuss mechanisms by which the governance-preserving attributes of DML frameworks can be supplemented to render them fully privacy-preserving, verifiable, and secure.
\end{itemize}

\section{Governance}
In its core, data governance is a framework that defines the way data should be handled from the perspectives of \textit{access}, \textit{ownership} and \textit{auditability} \cite{eryurek2021data}. For the purposes of distributed algorithmic processing, we consider auditability (that is, the ability of being inspected for e.g. quality) as a process that is either performed locally by the data owner on their own data or, alternatively, by the central aggregator on the updates submitted by the members of the consortium. We discuss auditability in the \textit{Accountability} Section. 

Central to exercising data governance is the ability to control data \textit{locality} and data \textit{accessibility}. In frameworks such as FL, control over data locality is maintained through local model training and subsequent update sharing, meaning that the data does not leave its owner owner and only the results of the algorithmic processing of this data are shared with the rest of the federation.

Nevertheless, DML allows (indirectly) accessing the data (or, more precisely, computational results derived from it), to the members of the federation. To prevent unwarranted or unauthorised access, security measures must be put in place, which are discussed below. In this regard, governance over data is maintained in DML paradigms such as FL, provided a suitable framework for auditing the input data is in place. However, even though appropriately audited data remains local and is only accessible through the collaborative learning protocol, the protection of privacy and secrecy are contingent on additional techniques being utilised. 

\section{Privacy and secrecy}
In this section we consider the fundamental differences between the concepts of \textit{secrecy} and \textit{privacy}, which are often conflated. 

Privacy is the ability to control how much can be learned from the data about an individual. In the context of ST, there exist two separate privacy-related concepts: input and output privacy. Input privacy controls the extent to which the input data is visible and accessible to other actors, whereas output privacy is concerned with how much can be learnt from the data itself. The former can be achieved through strategies that maintain the secrecy of the computation, whereby secrecy implies that the sensitive data itself cannot be seen by anyone other than the data owner, concealing it from other parties. While decentralised protocols such as FL or SL allow data owners to enforce governance over their data and prevent unwarranted access to the training data directly, they do not provide the contributors with any mechanisms to control what can be \textit{inferred} about that data when it is used for model training (output privacy). Input and output privacy pursue two complementary goals: secrecy prevents data from being usable when accessed inappropriately while privacy reduces the amount of information that can be extracted from the data about the individual while not wholly preventing drawing insights from the data otherwise.

The secrecy of datasets or of the computational process itself is often maintained through the application of formal cryptographic protocols. For example, individual contributions can be encrypted to conceal them from other participants during training. One such solution, namely homomorphic encryption (HE) \cite{gilad2016cryptonets,hesamifard2017cryptodl} allows the federation to run the entire training procedure on encrypted data and only decrypting the end result of the computation. Another solution, namely secure multi-party computation (SMPC) \cite{rouhani2018deepsecure, mohassel2017secureml} offers the federation an option of masking their contributions through splitting them into a number of encrypted \textit{shares}, which, unless aggregated with the agreement of a quorum of participants (sometimes all participants), do not allow another client to learn anything about the model update. This fact also renders SMPC an attractive option for distributed data governance, where only a given number of parties are able to reveal the data that is shared between them.
Most DML algorithms can be augmented with either of these mechanisms to provide the data owners with guarantees of secrecy, and, as a consequence, of input privacy. However, it is important to outline that typically such mechanisms usually come with a performance overhead that can affect the training time and/or the utility of the final model \cite{kaissis2021end,mohassel2017secureml}.

In contrast, enforcing output privacy requires limiting the information that can be derived about an individual from the data used for model training. While DML can guarantee data governance, it does not in itself offer \textit{any} meaningful privacy guarantees to the data owners. Therefore, the over-reliance on DML protocols alone, and their further adoption without the inclusion of formal privacy-preserving mechanisms must be viewed critically. Differential privacy (DP) \cite{dwork2006calibrating}, which allows to objectively quantify and bound the amount of information that can be inferred from the training data, has established itself as the gold standard of formal privacy protection. DP allows data owners to perform model training while offering the individuals whose data is used to train the model, a quantifiable privacy guarantee in the form of a \textit{privacy budget}. This privacy budget can be thought of as \textit{fungible}, because it can be expended through model training. After an individual's privacy budget is exhausted, no further interaction with their data is permitted. The combination of DML with DP and cryptographic tools can equip the federation with the means of ascertaining governance as well as input and output privacy. 
\section{Security}
In addition to the concepts of privacy and security outlined above, we additionally need to consider \textit{security}. In general, security is a property of a protocol or system where the protocol or system \textit{as-a-whole} cannot be threatened by an adversarial actor. Security is thus not just complementary to privacy and secrecy, but a prerequisite for the design of any private and secure system. For instance, the physical security of the buildings in which computational equipment is housed is paramount to ensuring that these systems cannot be tampered with or destroyed. Moreover, a series of software and hardware measures are employed to safeguard the security of the learning protocol. For example, technologies such as transport layer security (TLS) are deployed to encrypt the data \textit{in transit} and protect it against adversaries who could intercept data packets and obtain potentially sensitive information (so-called \textit{man-in-the-middle} attacks). In general, the security of the system concerns \textit{all} factors that can be exploited by adversaries that are covered in the ST framework and beyond. Issues such as physical unwarranted access to the data owner cannot be mitigated regardless of the privacy measures deployed, as these are covered under the notion of physical security. Similarly, issues such as attacks that are not exploiting/targeting the DML models directly (e.g. malicious hackers that attempt to steal the dataset through exploitation of the site itself rather than the trained model) do not lie within the scope of the ST framework. 
\section{Verification}
In order to verify that the model is trained well on the underlying learning task, the federation needs to establish a method to ascertain that the data of the federation was used faithfully and the training protocol was not subverted by any party. Currently, no published DML implementation support this notion of output verification, as it comes in direct conflict with the notions of privacy described above, as such verification could entail inspection of the training protocol and --as a result-- of sensitive data supplied by individual participants. One solution that can be employed for this task can build upon techniques from the verifiable computing (VC) domain to enable the actors of the federation to certify the results they compute. VC encompasses methods that can be used to attest that a given computation's result was produced by a specific instance of an algorithm, matching its dependencies to the very bit, as pre-agreed upon between the \textit{prover}, who runs the computation, and the \textit{verifier}, who seeks to obtain guarantees of integrity of the computation, i.e. that the result of the computation was no malformed and represents an honest processing of inputs by the algorithm.

There exist two main strategies of implementations of VC in the context of ML: one involves special hardware embodied as trusted execution environment (TEE) \cite{chen2020training,lee2020keystone} and relies on the hardware properties in conjunction with a mechanism termed remote attestation (RA) \cite{ali2018trust} to fulfil the verifiable trait of the execution. We note that TEEs (also termed \textit{secure enclaves}) can also be used for end-to-end encrypted computations, thus fulfilling the role of an input privacy mechanism.
The second form of VC that has gained traction over the past few years is entirely software-based and relies on cryptographic proofs. A popular form of cryptographic proof, namely \textit{zero knowledge succinct non-interactive arguments of knowledge} (zk-SNARKs) \cite{cryptoeprint:2014:580}, gained significant interest from the research community and has been previously leveraged in the context of neural network verification \cite{lee2020vcnn, weng2021mystique}. Augmented with these mechanisms, DML is able to guarantee that the computation was performed faithfully.

We note that --although verification of correctness is possible with these techniques-- the verification of data \textit{quality} is much more complex. For instance, a number of measures may be employed to gauge data quality, such as the reduction in model uncertainty obtained through each individual data sample. Moreover, members of the federation may agree upon metrics other than data quality (e.g. the speed with which a computational result is returned), to determine participant reimbursement.   

\begin{table*}[h!]
\centering
\resizebox{0.9\textwidth}{!}{%
\begin{tabular}{@{}ll@{}}
\toprule
\multicolumn{1}{c}{\textbf{Term}}             & \multicolumn{1}{c}{\textbf{Definition}}                                                                                                     \\ \midrule
\multicolumn{1}{l}{\textbf{Governance}}     & \multicolumn{1}{l}{Framework that defines the way data should be handled from the perspectives of access, ownership and auditability.}     \\ \midrule
\multicolumn{1}{l}{\textbf{Privacy}}        & \multicolumn{1}{l}{The ability to control how much can be learned from the data about an individual.}                                      \\ \midrule
\multicolumn{1}{l}{\textbf{Secrecy}}        & \multicolumn{1}{l}{Trait of the computation that implies that the sensitive data cannot be seen by anyone other than the data owner.}      \\ \midrule
\multicolumn{1}{l}{\textbf{Security}}       & \multicolumn{1}{l}{Property of a protocol or system where the protocol or system as a whole cannot be threatened by an adversarial actor.} \\ \midrule
\multicolumn{1}{l}{\textbf{Accountability}} & \multicolumn{1}{l}{The ability to track the source of the computation's results}                                                               \\ \midrule
\multicolumn{1}{l}{\textbf{Verifiability}}  & \multicolumn{1}{l}{The capability to guarantee an honest processing of the input data and the integrity of the computation}                \\ \bottomrule
\end{tabular} %
}
\caption{Summary of the relevant definitions}
\label{tab:defs}
\end{table*}
\section{Accountability}
\label{sec:accountability}
In a setting where several parties collaborate to solve a learning task, one cannot easily anticipate the intentions of each individual participant, as some actors might be actively attempting to subvert the training protocol. As a result, attacks on the utility of the resulting model are possible in decentralised learning, such as model poisoning \cite{fang2020local, yang2017generative} or backdoor attacks \cite{bagdasaryan2020backdoor, bagdasaryan2020blind}. Thus, paradigms such as FL require substantial augmentations to mitigate potential adversarial influence from being incorporated into the jointly trained model. This influence can take multiple forms ranging from colluding with other participants to submitting malformed updates or not submitting model updates at all, causing the training procedure to halt indefinitely. The ability to track the source of malicious interference is therefore a required component of trustworthy DML protocols. Moreover, as described above, it may also be required to track properties of the data pertaining to its quality, or about the computation, such as its speed or the result of a verification workflow.

Thus, certain DML implementations \cite{warnat2021swarm}, propose the utilisation of a permissioned blockchain to track the contributions of each individual data owner, discouraging them from submitting intentionally malformed model updates. Concretely, model aggregation and the selection of the aggregation server for the round (so-called \textit{leader}) occur through the execution of blockchain-backed smart contracts.

In general, the reliance on blockchains in DML can thus allow the federation to obtain a decentralised, immutable transcript of individual contributions to the training protocol. This can include the information about how each individual's contribution influences the resulting model, the time it took them to produce each contribution etc. Such information is essential in the identification of malicious actors whose contributions (or deliberate lack thereof, which can result in protocol halting), affect the utility of the jointly trained model. The immutability of blockchains additionally prevents such actors from concealing their contributions. Moreover, DML protocols combining the notions of data governance and accountability within the same distributed learning system have been developed \cite{passerat2020blockchain}. 

While accountability approaches undoubtedly increase the overall trust level, their contribution must be weighed against the cost of developing and deploying such complex solutions. A potential solution for future generations of truly trustless (i.e. those that can function when 50\% of all parties are assumed to be malicious) DML systems would be to benefit from leveraging a public blockchain network such as the Ethereum main network. On such truly decentralised public infrastructure, no participant in the federation can tamper with the data stored nor with the execution of the smart contracts.

\section{Conclusion}
In this work, we study the vulnerabilities associated with the na\"ive utilisation of DML. We deduce that most DML protocols cannot be relied upon unless accompanied by additional mechanisms enhancing trust between participants. Unfortunately, much published literature implies that FL and other DML protocols provide privacy protections. It is paramount to note that all that is offered by DML itself is --at best-- a \textit{semblance} of privacy. As a result, promoting the utilisation of DML without any formal measures of input or output privacy protection in place can lead to the disclosure of sensitive information, potentially causing irreparable damage. As evident from the comparison between the components of the ST framework and the current abilities of published DML systems, most frameworks can currently only satisfy a subset of ST requirements.
Moreover, we contend that the terms privacy, secrecy and security should not be used interchangeably, and that all are required for ensuring the trustworthiness of DML systems. As noted by \cite{carlini2021neuracrypt} \say{There is no room for error in privacy}, showing that the misinterpretation or entanglement of concepts can result in the violation of trust between parties of DML protocols. We summarise these (along with other) definitions in Table \ref{tab:defs}.

Finally, we propose the following recommendations that arise from our work:
\begin{enumerate}
    \item Future research in the field of trustworthy AI (particularly in the field of private ML) should agree upon and adhere to terminological guidelines. For instance, neither systems without formal privacy guarantees (e.g. FL) nor systems only offering input privacy (e.g. encryption) should be haphazardly termed \textit{privacy-preserving} (or similar).
    \item  Only systems adhering to all aforementioned principles should be termed \textit{trustworthy}, to avoid negative consequences associated with over-reliance on protocols that were not designed with fundamental privacy requirements in mind. Such systems should undergo external auditing in order to verify their correctness e.g. through the means of formal network certification \cite{lecuyer2019certified},
    \item Most existing DML solutions already offer promising foundations for emergence of private reliable ML systems, but they must be augmented with additional mechanisms in order to be able to guarantee both the privacy of the participants and the robustness of the jointly trained models. We emphasise the importance of education pertaining to these systems, both for experts and laypeople, in this regard.
\end{enumerate}
\bibliography{references.bib}
\end{document}